\documentclass[10pt]{article}
\author{Nicolas Valot$^1$, Louis Fabre$^1$, Benjamin Lesage$^2$, Ammar Mechouche$^1$, Claire Pagetti$^2$\\
$^1$  Airbus Helicopters, $^2$ ONERA}

\usepackage[disable]{todonotes}

\usepackage{graphicx} 
\usepackage[colorlinks=true, allcolors=blue]{hyperref}
\usepackage[top=2cm,bottom=2cm,left=1.5cm,right=1.5cm,marginparwidth=1.75cm,]{geometry}
\usepackage{comment}
\usepackage{tikz}
\usepackage{tikz-uml}
\usepackage[pdf]{graphviz}
\usepackage{listings}
\usepackage{xcolor}
\usepackage{caption}
\usepackage{xspace}
\usepackage{amsfonts}
\usepackage{amssymb}
\usepackage{amsmath}
\usepackage{import}
\usepackage[inline]{enumitem}
\usepackage{subcaption}
\usepackage{booktabs}
\usetikzlibrary{quotes,arrows.meta}
\usetikzlibrary{positioning}

\usepackage{Ball}
\usepackage{Box}

\usetikzlibrary{positioning}
\usetikzlibrary {shapes.misc}
\usetikzlibrary{shapes.geometric}
\usetikzlibrary{calc}  
\tikzset{cross/.style={cross out, draw=black, minimum size=2*(#1-\pgflinewidth), inner sep=0pt, outer sep=0pt},cross/.default={1pt}}
\usepackage{graphicx}
\graphicspath{ {./figures/} }
\setuptodonotes{fancyline, color=blue!30}
\presetkeys{todonotes}{inline}{}

\lstdefinestyle{mystyle}{
    backgroundcolor=\color{backcolour},   
    commentstyle=\color{codegreen},
    keywordstyle=\color{magenta},
    numberstyle=\tiny\color{codegray},
    stringstyle=\color{codepurple},
    basicstyle=\ttfamily\footnotesize,
    breakatwhitespace=false,         
    breaklines=true,                 
    captionpos=b,                    
    keepspaces=true,                 
    numbers=left,                    
    numbersep=5pt,                  
    showspaces=false,                
    showstringspaces=false,
    showtabs=false,                  
    tabsize=2
}
\definecolor{codegreen}{rgb}{0,0.6,0}
\definecolor{codegray}{rgb}{0.5,0.5,0.5}
\definecolor{codepurple}{rgb}{0.58,0,0.82}
\definecolor{backcolour}{rgb}{0.95,0.95,0.92}

\newcommand{\onnx}{{\sc ONNX}\xspace}
\newcommand{\ort}{{\sc ORT}\xspace}
\newcommand{\keras}{{\sc Keras}\xspace}

\newcommand{\arp}{{ED-324}\xspace}
\newcommand{\arpshort}{{\arp}\xspace}
\newcommand{\lenet}{{LeNet-5}\xspace}
\newcommand{\doseventi}{{DO-178/ED-12}\xspace}
\newcommand{\ieee}{{IEEE-754}\xspace}
\DeclareMathOperator{\var}{Var}
\DeclareMathOperator{\bias}{Bias}
\DeclareMathOperator{\mse}{MSE}
\DeclareMathOperator{\mae}{MAE}
\DeclareMathOperator{\mape}{MAPE}
\DeclareMathOperator{\rsq}{R^2}
\DeclareMathOperator{\evs}{EVS}
\DeclareMathOperator{\argmax}{argmax}
\DeclareMathOperator{\iou}{IoU}
\DeclareMathOperator{\mAP}{AP}
\DeclareMathOperator{\topone}{Top-1}

\newcommand{\implmdl}{{TIM}\xspace}
\newcommand{\trainmdl}{{TFM}\xspace}

\newtheorem{definition}{Definition}
\newtheorem{notation}{Notation}
\newtheorem{example}{Example}

\newtheorem{objective}{Objective}

 \newcommand{\ucf}[1]{\texttt{#1}}
\newcommand{\usecase}[5]{\ucf{#1}-\ucf{#2}-\ucf{#3}-\ucf{#4}-\ucf{#5}}
 \newcommand{\uclstm}{\ucf{lstm}}
 \newcommand{\uclinear}{\ucf{linear}}

\title{Implementation of airborne ML models with semantics preservation}
\date{}

\begin{document}

\maketitle

\pagestyle{plain}

\begin{abstract}
  Machine Learning (ML) may offer new capabilities
  in airborne systems.
However, as any piece of airborne systems, ML-based
systems will be required to guarantee their safe operation.
Thus, their development will have to be demonstrated to be compliant with the adequate guidance.
So far, the European Union Aviation Safety Agency (EASA) has published a concept paper and an EUROCAE/SAE group is
  preparing \arp. 
Both approaches delineate high-level objectives to confirm the ML model achieves its intended function and maintains training performance in the target environment.

The paper aims to clarify the difference between an ML model and its corresponding unambiguous description, referred
  to as the Machine Learning Model Description (MLMD).
It then refines the essential notion of semantics preservation to ensure
the accurate replication of the model.
We apply our contributions to several industrial use cases
  to build and compare several target models.
\end{abstract}
\todo{Remember to uncomment the command \textbackslash usepackage[disable]\{todonotes\} at the top of the main.tex to hide comments}
\todo{Remember to comment the command \textbackslash pagestyle\{plain\} at the top of the main.tex to fit in the conference template.}

\section{Introduction}
\label{sec:introduction}

Machine Learning (ML) has gained increased consideration even in airborne avionics systems.
However, the introduction of ML algorithms in avionic embedded systems challenges 
the established practices of the development assurance industry.
Thus, it has led to the emergence of new development assurance processes, as outlined in the 
EASA guidance \cite{Easaconcept}
and the yet to be published draft of the \arp, with publicly available material \cite{gabreau:hal-WG,kaakai2023datacentric,kaakai2022toward}.
Both documents are limited to the design assurance levels related to the least critical failure conditions: Major and Minor.

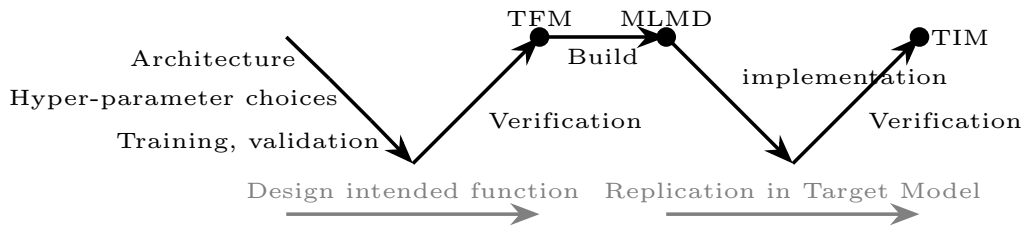
\begin{figure}[hbt]
	\vspace{-1em}
	\centering
	\resizebox{0.75\textwidth}{!}{%
	\begin{tikzpicture}
	\tiny
	\draw[black, thick]  (-2,0) to  node[left]{Architecture}  (-1.66,-0.33);
	\draw[black, thick]  (-1.66,-0.33) to  node[left]{Hyper-parameter choices} (-1.33,-0.66);
	\draw[black, thick][-Stealth]  (-1.33,-0.66) to node[left]{Training, validation} (-1,-1);
	\draw[black, thick][Stealth-]  (0,0) to ["Verification"] (-1,-1);
	
	\draw[black, thick][-Stealth]  (0,0) to node[below]{Build}  (1,0);
	
	\draw[black, thick][-Stealth]  (1,0) to ["implementation"]  (2,-1);
	\draw[black, thick][Stealth-] (3,0) to ["Verification"] (2,-1);
	\draw[gray, thick][-Stealth] (-2,-1.4) to ["Design intended function"] (0,-1.4);
	\draw[gray, thick][-Stealth] (1,-1.4) to ["Replication in Target Model"] (3,-1.4) ;
	\filldraw[black] (0,0) circle (2pt) node[above]{\trainmdl};
	\filldraw[black] (1,0) circle (2pt) node[above]{MLMD};
	\filldraw[black] (3,0) circle (2pt) node[right]{TIM};
	\end{tikzpicture}
	}%
	\caption{\arp W-shape Development Life Cycle\label{fig:MLlifecycle}}
\end{figure}

\textbf{Guidance approach.}
These documents address supervised, off-line trained, ML models and promote a W-shaped development life-cycle (outlined in Figure~\ref{fig:MLlifecycle}).
It  consists roughly of two main phases: 
\begin{enumerate*}
\item the design of the intended function~(first V-cycle),
and \item its replication in the Target Model~(second V-cycle).
\end{enumerate*}
 The Target Model~(\implmdl) captures both representation of the implemented ML model, and the
 target environment, i.e.\ the hardware and software platform and its configuration used to execute the model.
The second V-cycle
first implements the Machine Learning Model Description (MLMD) into a \implmdl, while ensuring the
 semantics
 preservation of the off-line trained model.
 The verification then assesses the correct replication of the TFM by the implementation.


As guidance, the new development assurance processes exhibit some leeway in their interpretation, and they do not aim to be prescriptive of specific tools or methods.
This led to a recent, and growing interest in providing techniques to support the certification of ML-based systems across the W-shaped life-cycle~\cite{TowardsDesignAssuranceLevelC, Wasson2024DeobfuscatingML, Christensen2024TowardsCA, Elboher2024RobustnessAO}.
Our approach 
complements those works with a consistent method to comply with the certification objectives related to the replication phase.


\textbf{Focus of the paper.}
This paper focuses on the construction of the MLMD
as the \emph{bridge} between the two
phases.
The Training Framework Model (\trainmdl) results from off-line training and verification.
It is expressed using the internal representation of a training framework.
By contrast, the MLMD is a non-volatile and semantically-defined ML model, which is independent of training concerns, i.e.\ the MLMD does not contain learning rate or loss function.
To build the MLMD, it is essential
to first identify and formalise the TFM behaviour as well as the properties that must be maintained during implementation.
The verification at the end of the first V phase
aims to determine whether the ML model
satisfies  requirements and properties such as
stability, generalisation, performance and robustness. 
Therefore, the implementation is expected to accurately encode the ML model mathematical
operations~(e.g.\ convolution or pooling) while preserving all properties fulfilled by the \trainmdl.

Several verification strategies can be employed.
First, the applicant replicates all the verification steps performed at the end
of the first V phase on the \implmdl.
 If any property is not preserved, they must develop a process to correct the implementation.
 Second, the applicant may choose their preferred training framework (e.g., Keras with PyTorch), fully reverse the \trainmdl-associated code implementation, and re-implement the executable-level semantics identically.
This procedure is highly costly and only applicable to a particular version of the training framework, which evolves regularly.
We propose an alternative to these two extreme approaches.


\textbf{Contributions.}
 We propose a definition of semantics preservation supporting the \arpshort objectives and the \textit{Anticipated MOC-SA-01-2} of EASA concept paper \cite{Easaconcept}.
 It ensures both the~\trainmdl and the~\implmdl satisfy the same properties,
 as verified by the \trainmdl verification metrics.
 In effect,
 the applicant formalises
 \begin{enumerate*}
 \item the properties to be kept,
 \item the behavioural discrepancies between the~\trainmdl and the~\implmdl,
 and \item demonstrates that the properties are preserved under certain conditions.
 \end{enumerate*}
We outline a method to assess whether a~\implmdl preserves said properties.
 We export the \trainmdl in an intermediate format.
There are numerous formats available for exchanging ML models between different training or deployment frameworks (Tensorflow-lite, NNEF, StableHLO, OpenVINO, the Open Neural
Network eXchange~(\onnx)).
In this work, we consider \onnx-based MLMD as this format is
widely used in the ML domain and embedded domain.
Since \trainmdl can be very large, manual
production of a MLMD is not sustainable and \emph{exporters} should be considered.
We evaluate our approach on industrial use-cases, using state-of-the-art tooling to generate~\implmdl.
 Our evaluation shows that the resulting models can satisfy the same properties as the \trainmdl, even at reduced numerical precision or on an embedded platform.

In the end, the paper answers the question: can a ML Model be demonstrated as exactly replicated on the target system?
Which approaches could be applied for that?

 \textbf{Outline.}
 The paper is organised as follows.
 Section~\ref{sec:arp} introduces the description of an ML model, in relation to the \arpshort objectives.
 It provides the background required to define our proposal to verify the semantics preservation from~\trainmdl to~\implmdl.
 We discuss the application of our method in Section~\ref{sec:applicability}, considering how model transformations impact the~\implmdl lifecycle.
To support our evaluation, Section~\ref{sec:setup} introduces the considered \trainmdl case studies and methods to build~\implmdl.
 The verification of both~\trainmdl and~\implmdl is performed in Section~\ref{sec:results}.
Finally, we consider how our approach fits with existing work in  Section~\ref{sec:related-work}, before concluding in Section~\ref{sec:conclusion}.

\section{How to describe a ML model}
\label{sec:arp}

\arp  defines a design assurance process to conceive and
develop ML-based systems.
One important expectation  is to ensure that the \implmdl reproduces the \trainmdl behaviour and the properties observed at the end of the design,
and to facilitate its verification.
The approach proposed by the standard is to introduce an intermediate description, the MLMD,
between the two V cycles.
The purpose of  MLMD is to describe  ML models in an unambiguous way,
so that the implementation can start with a complete specification.

\subsection{Understanding which properties to preserve}
\begin{figure}[hbt]
  \centering
  \resizebox{0.5\linewidth}{!}{%
	\begin{tikzpicture}
	\node [rectangle, minimum height = .5cm,draw] (valset) at (0,0){\begin{tabular}{l}Test\\ dataset\end{tabular}};
	\node [rectangle, minimum height = .5cm,draw] (gt) at (3,-1){\begin{tabular}{l}Ground\\truth \end{tabular}};
	\node [rectangle, minimum height = .5cm,draw] (pred) at (3,1){\begin{tabular}{l}\trainmdl\\ prediction\end{tabular}};
	\node [rectangle, minimum height = .5cm,draw] (val) at (6.25,0){\begin{tabular}{l}Verification with\\metrics and bounds\end{tabular}};

	\draw [-Latex] (valset) |-   (pred.west);
	\draw [-Latex] (valset) |-   (gt.west);
	\draw [-Latex] (gt) -|   (val);
	\draw [-Latex] (pred) -|   (val);
	\end{tikzpicture}
        }
	\caption{\trainmdl verification with a set of metrics\label{fig:tfm_validation}}
\end{figure}
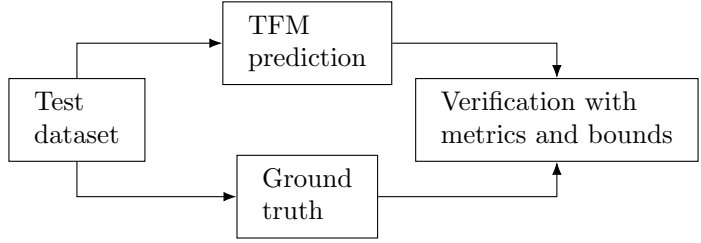

First, let us briefly outline what is done during the verification of the first V.
A ML model is expected to fulfill some requirements and properties such as 
stability, generalisation, performance, robustness. 
Practically, 
several metrics are used to check whether those properties are met,
and to identify the conditions under which those properties are fulfilled~(e.g.\ range of input data).
Figure~\ref{fig:tfm_validation} highlights
this verification by the data scientists at the end
of the first V.
They consider a set of methods and metrics,
and
they check that the error between the prediction
and the ground truth,
 for a (combination of) \emph{metric},
 fits within acceptable bounds for the test dataset.

\begin{notation}
 \label{not-d-f1}
  \it
  Let us consider a dataset composed of $n$ input samples
  $\mathcal{D}= \{x_i\}_{i\in[1,n]}$ with $x_i \in \mathbb{R}^p$.
  Let us denote by $f:\mathbb{R}^p \rightarrow \mathbb{R}^m$ the unknown function  to be fitted by the ML model.
  Thus,
  $f(x_i)$ denotes the ground truth associated to the input sample $x_i$. 
  Let us denote by $\hat f_1:\mathbb{R}^p \rightarrow \mathbb{R}^m$
  the function realised by the \trainmdl.
\end{notation}

\begin{example}[Metric $L_\infty$]
  \label{ex-metric-linf}
 A very simple metric is given by  $L_\infty$: the idea is to check for all $x_i\in \mathcal{D}$
 that the prediction of the \trainmdl is close enough to the ground truth, i.e.  $L_\infty(f,\hat f_1)= \emph{sup}_i|f(x_i) - \hat f_1(x_i)|\leq R_{L_\infty}$
 where $R_{L_\infty} \in \mathbb{R}_+$ is an acceptable bound set by the designer.
\end{example}

\begin{example}[Metric Bias]
  \label{ex-metric-bias}
 A second widespread metric is the Bias, often used for regression tasks, with
   $\bias(f,\hat f_1)= \frac{1}{n} \sum^n \left(\hat f_1(x_i) -  f(x_i)\right)$.
The associated \trainmdl verification is to check whether  $|\bias|\leq R_{\bias}$
 where $R_{\bias}\in \mathbb{R}_+$ is an acceptable bound set by the designer.
\end{example}

 \begin{example}[Metric MAE]   
Lastly,  if
\emph{generalisation} is one of the  properties,
the associated metric is the Mean Absolute Error (MAE),
where
$\mae(f,\hat f_1)= \frac{1}{n} \sum^n |\hat f_1(x_i) - f(x_i)|$.
Again, the condition to be checked is  $\mae\leq R_{\mae}$
 where $R_{\mae}\in \mathbb{R}_+$ is an acceptable bound set by the designer.

\end{example}

\subsection{\arp objectives related to MLMD}
\label{sec:arp-mlmd}
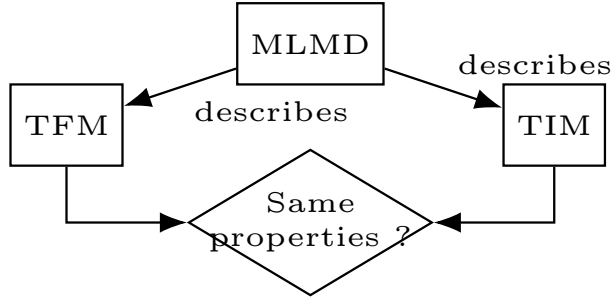
\begin{figure}[hbt]
	\tiny
	\centering
	\resizebox{0.45\textwidth}{!}{%
	\begin{tikzpicture}
	\node [rectangle, minimum height = .5cm,draw] (MLMD) at (0,0){MLMD};
	\node [rectangle, minimum height = .5cm,draw] (TFM) at (-1.5,-0.5){\trainmdl};
	\node [rectangle, minimum height = .5cm,draw] (Target) at (1.5,-0.5){\implmdl};
	\node [diamond, aspect = 2, minimum height = .9cm, minimum width = 1.5cm, draw] (props) at (0,-1.1){};
        \node (propstext) at (0,-1.1){\begin{tabular}{c}Same\\properties ? \end{tabular}};
       
	\draw [-Latex] (MLMD) to  ["describes"] (Target);
	\draw [-Latex] (MLMD) to   ["describes"] (TFM);
	\draw [-Latex] (TFM) |-   (props);
	\draw [-Latex] (Target) |-   (props);
	\end{tikzpicture}
	}%
	\caption{MLMD to guarantee ML Model properties\label{fig:properties}}
\end{figure}

The second V of the W-shape aims at preserving
the properties satisfied by the \trainmdl (see previous section).
For this purpose, \arpshort relies on the definition of the MLMD
as a bridge between ML model design (\trainmdl) and ML model implementation,
that we call the \emph{Target Model}~(\implmdl).
The purpose of the MLMD is to ensure that the \implmdl reproduces
the \trainmdl observed behaviour, with the intent that they both satisfy the same properties.
This is highlighted in Figure~\ref{fig:properties}.

To achieve this, the standard defines several objectives, and
we refine those
related to the construction of the MLMD\@.
The  ML Model Description (MLMD), as its name suggests, is an
artifact that maintains a record of both the ML model design
and its final configuration. 
A ML model is specified as a directed acyclic graph,
the nodes of which are the operators
and the edge the data-flow between operators.
The first refined objective states that
these computations
are thoroughly captured by the MLMD, and is related to Objective IMP-04 of \cite{Easaconcept}.
\begin{objective}[Capturing the elementary functions and composition rules]
  \label{obj:semantic}
  \it
  The function associated to each operator must be defined in a non ambiguous way, and the rules for combining
  them, to produce the function carried out by the neural network, must also be provided.
\end{objective}

The second refined objective related to Objective SA-01 and IMP-06 of \cite{Easaconcept}
requires the~\implmdl to \emph{accurately} reproduce
the \trainmdl behaviour
and this is expressed by \emph{semantics preservation}.
\begin{objective}[Semantics preservation]
   \label{obj:replication}
   \it
   The MLMD must come with an unambiguous semantics
   which must be preserved in the~\implmdl.
\end{objective}

Thanks to these objectives, the verification activities done at the end
of the design phase do not have to be redone in the~\implmdl.
However, those objectives are not prescriptive to allow applicant for choosing
their own (internal industrial) process development
and as such, they are open to interpretation.
Our purpose is to clarify our understanding of their meaning, propose an interpretation
with an associated method to achieve them,
and bring some formal arguments towards correctness.

\subsection{Semantics preservation to maintain ML properties}
\label{sec:arp-preservation}

Let us start with refining the meaning and purpose of those two objectives.
Objective \ref{obj:semantic} concerns the description of the
ML model as a directed acyclic graph,
 capturing the dataflow and operations in the ML model.
One natural (and minimal) way to describe such model is a \emph{mathematical representation}.
Each operator
is a function over $\mathbb{R}$ or $\mathbb{Z}$
defined by mathematical expressions with basic operators~(e.g. $+,-,\times,/,\exp$).
\cite{silva_acetone_2022} provides such a formal description for simple neural networks.
The notion of semantics preservation
of Objective~\ref{obj:replication}  can encompass matching the pure numerical output of a ML model
or a more context-dependent definition, such as the classification it produces.
Let us complete Notation~\ref{not-d-f1}.
\begin{notation}
  \label{not-f2}
  \it
  Let us denote by $\hat f_2:\mathbb{R}^p \rightarrow \mathbb{R}^m$
  the function realised by the~\implmdl.
\end{notation}

\begin{figure}[hbt]
  \centering
  \begin{tikzpicture}
    \filldraw[color=gray!60, fill=gray!5, very thick](0.4,-0.3) circle (0.35);
    \draw (0,0) node[cross]{ };
    \draw (0.4,-0.3) circle (1pt);
    \node  [regular polygon, regular polygon sides = 3,draw,inner sep=0.5pt] at (0.4,-0.6){};
    \node  at (-1,0.6){$f(x)$};
    \node  at (-1,0){$\hat f_1(x)$};
    \node  at (-1,-.8){$\hat f_2(x)$};
    \node  at (-2.1,0.1){$x$};
    \draw[dashed] (-2,0) edge[bend left, -Latex]  (-0.05,0.05);
    \draw (-2,0) edge[bend right, -Latex]  (0.35,-0.6);
    \draw  (-2,0)  edge [out=-20, in=190, -Latex]  (0.35,-0.3);

    \filldraw[color=gray!60, fill=gray!5, very thick](3,0) circle (1);
     \draw (3,0) node[cross]{ };
    \draw[dashed] (0.04,0.04) edge[bend left, -Latex]  (2.96,0.05);
    \node  at (1.5,0.8){$M(f,f)$};
    \draw (0.4,-0.6) edge[bend right, -Latex]  (2.75,-0.6);
    \node  [regular polygon, regular polygon sides = 3,draw,inner sep=0.5pt] at (2.8,-0.6){};
    \node  at (1.5,-1.2){$M(f,\hat f_2)$};
    \draw  (0.42,-0.3)  edge [out=-20, in=190, -Latex]  (2.85,-0.3); 
    \draw (2.9,-0.3) circle (1pt);
    \node  at (1.5,-.2){$M(f,\hat f_1)$};
    \draw(3,0)--(4,0);
    \node  at (3.5,.2){$R_M$};
    \node  at (3.2,-0.5){\footnotesize$g_M$};
    \node  at (0.85,0.1){\footnotesize$\varepsilon_M$};
    \draw(0.4,-0.3)--(0.55,0);
    
  \end{tikzpicture}
  \caption{Semantic preservation in the~\implmdl\label{fig:replication}}
\end{figure}
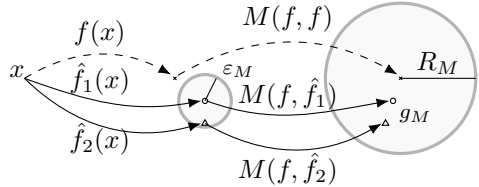

Figure~\ref{fig:replication} represents the different predictions ($f$, $\hat f_1$, $\hat f_2$)
and their satisfaction of a property according to a metric $M$ and a bound $R_M$.
For the sake of simplicity and 2D representativeness,
we suppose that $f:\mathbb{R}^p \rightarrow \mathbb{R}$.

The acceptable error requirement for  a given metric $M$
is depicted by the larger circle of radius $R_M$
 in Figure~\ref{fig:replication}.
 Intuitively, 
predictions from the~\implmdl
are subject to the same constraints as the \trainmdl to preserve the semantics of the ML model.
There are some discrepancies between the predictions of the \trainmdl and the~\implmdl~\cite{jajal2024interoperability},
due to the different model representations or execution environments.
This is highlighted by the small circle centered by the \trainmdl prediction.

 \todo{BL: Pas certain de l'inclusion de la Ground Truth dans la definition, si on mentionne qu'on en a pas besoin pour notre vérification par ailleurs.}
\begin{definition}[Semantics preservation in the~\implmdl]
  \label{def:preservation}
  \textit
  The~\implmdl preserves the semantics if it implements MLMD and preserves the verification 
  against the ground truth.
  In fact,
  for any metric $M$ and their associated bounds $R_M$,
  $M(f,\hat f_1)$ is inside the big circle with $M$ computed over $x_i\in \mathcal{D}$
  $\implies$
$M(f,\hat f_2)$ is also inside the big circle.
\end{definition}

As suggested by Objective IMP-05 in \cite{Easaconcept}, \implmdl is to be implemented through \doseventi \cite{DO178} development life-cycle for which the metrics $M$ may be included in the High Level Requirements. 
We propose a method to assess whether the~\implmdl preserves the verified
ML model properties by comparing the predictions of the
\trainmdl and~\implmdl.

\subsection{Proposed approach to ensure semantic preservation}
\label{sec-proposed-approach}
The method relies on the provision by the model designer
of all metrics $M$ used during the verification and on
an acceptable bound $R_M$.
In this work, we assume that a
 metric is a function $M:\mathbb{R}^m\times \mathbb{R}^m \rightarrow \mathbb{R}$.
For all metric $M$, the implementation engineer
must derive an
error margin $\varepsilon_M$
and a positive
budget $g_M$ 
such that:
{\small
  \[
|M(f, \hat f_1)|\leq R_M - g_M \wedge L_\infty(\hat f_1, \hat f_2) \leq \varepsilon_M
\implies |M (f,f_2)| \leq R_M 
\]
}
For some metrics, the constraints over $M$ are to be greater than, i.e.
$|M(f, \hat f_1)|\geq R_M + g_M \wedge L_\infty(\hat f_1, \hat f_2)
\leq \varepsilon_M
\implies |M (f,f_2)| \geq R_M$.
In all cases,
if such error margin $\varepsilon_M$ can be computed for all metrics budget $g_M$,
then the~\implmdl preserves the semantic of the~\trainmdl.

 \todo{BL: L'exemple~\ref{ex-methodo_linf} est raide. La valeur de $\varepsilon$ n'est pas si triviale de la dernière inéquation, et celle de $g$ donne $R_\infty - g_\infty = L_\infty(f, f_1)$. D'où une condition simplifiée:
     $L_\infty(\hat f_1, \hat f_2) \leq R_\infty - L_\infty(f, f_1)
     \implies L_\infty (f,f_2) \leq R_\infty$. Il y a une condition implicite, parce que $g_M > 0$: $R_\infty - L_{\infty}(f, f^1) \geq 0$
 }

\begin{example}[$L_\infty$ metric]
  \label{ex-methodo_linf}
  Let us illustrate how to apply the approach when the metric is the $L_\infty$ (see example \ref{ex-metric-linf}).
  We have seen that  $L_\infty(f,\hat f_1)= \emph{sup}_i|f(x_i) - \hat f_1(x_i)|$.
 Therefore,  $L_\infty(f,\hat f_2)= \emph{sup}_i|f(x_i) - \hat f_2(x_i)|= \emph{sup}_i|f(x_i) - \hat f_1(x_i)+\hat f_1(x_i) - \hat f_2(x_i)|\leq L_\infty(f,\hat f_1) + L_\infty(\hat f_1,\hat f_2)$.
  Thus, there exist an acceptable error margin during implementation $\varepsilon_{\infty}=R_{L_\infty}-L_\infty(f,\hat f_1)$
  and a formula
  $g_{\infty}=\varepsilon_{\infty}$.
  Note that in the worst case, $\varepsilon_{\infty}=0$ and the implementation must reproduce identically the \trainmdl.
\end{example}

\begin{example}[Regression task and bias metric]
   \label{ex-methodo_bias}
   Let us apply the approach on the bias metric (see example \ref{ex-metric-bias}).
We have seen that 
$\bias_1 = \frac{1}{n} \sum^n \hat f_1(x_i) -  f(x_i)$.
Therefore,
$\bias_2 = \frac{1}{n} \sum^n \hat f_2(x_i) -  f(x_i)=  \frac{1}{n} \sum^n \hat f_2(x_i) - \hat f_1(x_i) +\hat f_1(x_i) - f(x_i) =  \frac{1}{n} \sum^n \hat f_2(x_i) - \hat f_1(x_i) + \bias^{(1)}$.
This entails $|\bias_2| \leq L_\infty (\hat f_1,\hat f_2) + |\bias^{(1)}|$.
Thus, there exist an acceptable error margin during implementation $\varepsilon_{\bias}=R_{\bias}-L_\infty(f,\hat f_1)$
  and a formula
  $g_{\bias}=\varepsilon_{\bias}$. 
\end{example}

With the previous examples, we see that for each metric, the implementation engineer must find some formula relating $M (f,\hat f_2)$ with  $M (f,\hat f_1)$. We will see in the next section
that this is possible for the usual metrics used by data scientist.
The second question is how to set $\varepsilon_M$ (or the small circle) to fit the budget $g_M$
if necessary.
The MLMD
can be given at different levels of abstraction.
We refer to this concept as 
\emph{semantics-level-replication} (or SL-replication).
We propose 4 levels SLx of semantics
inspired by~\cite{floyd1993assigning,hoare,denotational}, to address Objective IMP-07 of \cite{Easaconcept}.
\textbf{SL0 is the pure mathematical notation} mentioned before.
The last level SL3 is met when
the predictions of the training framework are strictly equal to the~\implmdl predictions.
This corresponds to the bit-accurate-replication (i.e. $\forall x_i\in \mathcal{D}, \hat f_1(x_i) = \hat f_2(x_i)$).
SL$k$ is a complementary specification of SL$k-1$ (it encapsulates SL$k-1$).
Thus, the higher $k$ is in
the SL$k$ at which \trainmdl and~\implmdl match, the smaller is the $\varepsilon_M$ (the small circle in Figure~\ref{fig:replication})
 leading to a looser constraint over $M(f, \hat f_1)\leq R_M - g_M$ but at the cost of reaching SL$k$.

\textbf{SL1 - Machine number representation based level:}
The ML model is specified using a mathematical description similar to the mathematical notation SL0,
with functions defined over
machine number representations, e.g.\ 64-bit floating point~(FP64) or 8-bit integer~(INT8).
The explicit behaviour of operations shall be specified for cases where their expected output in $\mathbb{R}$
is outside the range of the selected representation.
Indeed, operations over machine number representations might overflow due to the limited precision of the representation.
 As an example, the INT8 range is $[-128, 127]$, and $127 + 1$ results in an overflow.
We denote by $\mathbb{D}_b$ a $b$-bit encoded representation of domain $\mathbb{D}$, and $\mathbb{F}_b$ in particular denotes $b$-bit floating point numbers.
The §5 of the \ieee standard \cite{IEEE754} supports this level of semantics for operations on numbers in $\mathbb{F}_b$ for $b \in \{64,32,16\}$.

\textbf{SL2 - Operational semantic level:}
 The ML model specification should decompose tensor and matrices expressions into an explicit ordering of all elementary operations on scalar values, and the approximations used for all mathematical operations.
Indeed, $\mathbb{F}_b$ data types raise associativity concerns.
It may be that for some $(a,b,c) \in \mathbb{F}^3_b$ and $\textit{ op }\in\{+,\times\}$,
$(a \textit{ op } b) \textit{ op } c \neq a\textit{ op } (b\textit{ op }c)$.
In addition, mathematical functions ($\exp, \tanh, \sigma ...$) are usually specified by a polynomial, resulting in an ordered sequence of \textit{ op }.
Therefore, the order of all \textit{ op } computations is defined.
The specification is thus akin to three-address code~\cite{dragoonBook} where each instruction features at most three scalar operands, that is an assignment and a binary operator.
Some operation orderings may mitigate the rounding error,
as example using pair-wise summation, the Kahan algorithm \cite{kahan} or the usage of augmented arithmetic described in §9.5 of \ieee.

 \todo{BL: Je ne sais pas si SL3 doit aussi traiter le cas de précision réduite vis-à-vis de la spec SL1.}

\textbf{SL3 - Execution model level:}
The ML model specification maps expression operands and operations onto hardware resources.
Expression operands might be assigned in registers or memory.
Similarly, operations may be mapped to specialised or extended precision arithmetic units, such as fuse-multiply-add (FMA) for an $(x\times y + z)$ operation.
The decision to assign an operand to a register, or an operation to a specific unit might result in operations performed at extended precisions over the SL2 specification, thus impacting the rounding effects.
 At this level, the rounding policy (to nearest, towards zero, \ldots) is defined which has an effect on each addition, multiplication in $\mathbb{F}_b$.
The environment, and in particular the compilation tool-chain, play a crucial role in a SL3 specification.
As an example, register spilling occurs when the compiler has to push intermediate operands into memory, instead of registers.
The results of computation will be rounded when pushed to memory, while they would not if the computation remained in extended precision registers.
At source level (C language), one can enforce rounding using the \emph{volatile} qualifier on local variables which forces data to be pushed and fetched from memory.
The optimising compiler might similarly elect to map, or not, an operation onto the FMA.

\section{Applicability of the approach}
\label{sec:applicability}

In this section, we detail the approach proposed in Section \ref{sec-proposed-approach} on different metrics to show that it is always possible to construct
$\varepsilon_M$ and $g_M$.
We then focus on how to build an MLMD at a \emph{replication-semantics-level} SLx.
We choose to rely on
the \onnx format,
one of the most widely-supported by popular training frameworks in the ML community.
This makes \onnx a prime candidate as a MLMD format supporting
Objective~\ref{obj:semantic}.

\subsection{Constructing $g_M$}
Let us recall some ML metrics (§2 in \cite{mleap}). Our list is not exhaustive meaning that an applicant would have to extend them for any other metrics or any other dimensionality (e.g. if $M$ is not in $\mathbb{R}$).

\textbf{Regression task main metrics}
The main metrics are
the $L_\infty$,
Bias, Mean Absolute Error ($\mae$), Mean Square Error ($\mse$), Variance, Explained variance score (EVS), Coefficient of determination ($R^2$), Mean Absolute Percentage Error~($\mape$) metrics.
The case of $L_\infty$ (resp. $\bias$)
has been shown in Example~\ref{ex-methodo_linf} (resp. \ref{ex-methodo_bias}).
The others are shown in Figure~\ref{fig-metric-g}, the
equations are detailed in Section~\ref{sec:annex}.

\begin{figure}[hbt]
\begin{center}
\begin{tabular}{|l|l|l|}
  \hline
  Metric & TFM threshold & $g_M$\\
  \hline
  MAE &  $\leq R_M-g_M$ & $\varepsilon_{\mae}$\\
  MSE &  $\leq R_M-g_M$ &$\varepsilon_{\mse}^2 + 2\varepsilon_{\mse}|\bias^{(1)}|$\\
  Var &   $\leq R_M-g_M$ &$\varepsilon_{\var}^2 + 2\varepsilon_{\var}|\bias^{(1)}|$\\
  MAPE &  $\leq R_M-g_M$ &$\varepsilon_{\mape} (1 + \mape^{(1)})$\\
  EVS &  $\geq R_M+g_M$ &$\frac{\varepsilon_{\evs}^2 + 2\varepsilon_{\evs}|\bias^{(1)}|}{\var(f(x))}$\\
  $R^2$ &  $\geq R_M+g_M$ &$\frac{\varepsilon_{R^2}^2 + 2\varepsilon_{R^2}|\bias^{(1)}|}{\var(f(x))} $\\
  \hline
\end{tabular}
\end{center}
\caption{Formula $g_M$ for the main regression tasks metrics\label{fig-metric-g}}
\end{figure}

\textbf{Classification and Decision task metrics}
We can extend this method to classification tasks considering that $\hat f(x_i)$ are the logits, i.e.\ the numerical values associated with the classes, and preceding the \emph{argmax} function which selects the highest scoring class.
The metric for classification is $\topone$ accuracy score, which is the portion of the number of True Positive and True Negative in the number of samples:
\todo{BL: $f(x)$ devrait être une classe directement, non? Seuls les $\hat f$ sont définis par leur logits.}
$\topone(f,\hat f_2) = \frac{1}{n} \sum_i \left(\argmax(\hat f_2(x_i)) = cid(f(x_i))\right) \ge R_{\topone}$, where \emph{cid} is the identifier of the ground truth class.
\begin{align*}
\small
&\topone(f,\hat f_1) \ge R_{\topone} + g_{\topone} \\
&g_{\topone} = \topone(f,\hat f_1) - \min(\topone(f,\hat f_2))  \\
&g_{\topone}  = \topone(f,\hat f_1) - \min(\topone(f,\hat f_1 \pm \varepsilon_{\topone})) 
\end{align*}

To each value of $\varepsilon_{\topone}$ corresponds a value of $\topone$ and vice versa.
To find the maximum $\varepsilon_{\topone}$ to get a given minimal $\topone$ score (and $g_{\topone}$),
we need to sample values of $\varepsilon_{\topone}$ and compute the $\topone$ scores.
This builds a Lookup Table (LUT). We can invert the relationship by interpolating the LUT:
$\varepsilon_{\topone} = interp(LUT(\topone))$.
More generally, the Top-N score is the condition that $cid(f(x)) \in \argmax_N(\hat f_2(x))$, where $\argmax_N$ returns the $N$ highest $\hat f_2(x)$.


\textbf{Object detection task metrics}
Object detection models combine
regression and classification, computing respectively the coordinates of a bounding box and its class.
The commonly used object detector property is a combination of precision and recall summarized by the AP (Average Precision) indicator, which is based on True positive, False positive, and False negative detections \cite{metrics}.
The Intersection over Union ($\iou$) is the Jaccard index metric related to the predicted bounding box and the ground truth bounding box.
Let $a$ be the sample input, and a detection bounding box $\hat f_2(a)$ in the~\implmdl.
The False Negative counter is incremented for each bounding box in  $f(a)$ with no associated prediction $\hat f_2(a)$.
If there is a bounding box in $f(a)$ such that $\iou_{(2)} > threshold$ (threshold is usually chosen above 50\%) and $cid(\hat f_2(a))$ is identical to $cid(f(a))$,
the True Positive counter is incremented.
Otherwise, the False Positive counter is incremented.

\begin{figure}[hbt]
	\centering
	\includegraphics[scale=0.5]{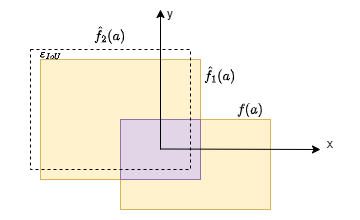}
	\caption{$\min \iou metric$}
	\label{fig:boundingbox}
\end{figure}

In the figure \ref{fig:boundingbox}, the bottom right rectangle is the ground truth bounding box $f(a)$, the top
left rectangle is the \trainmdl prediction $\hat f_1(a)$, and the dashed rectangle is the prediction in the~\implmdl $\hat f_2(a)$ determined by the $\pm \varepsilon_{\iou}$ added to the $\hat f_1(a)$ corners coordinates, which corresponds to the minimal $\iou$.
The $\iou$ is defined by the intersection (in purple) of surfaces over their union (combining purple and yellow).

\todo{BL: Clairement ici on ne colle pas trop à $g_M$ et $\varepsilon_M$. $M$ c'est $mAP$ ou $\iou$?
NV: M c'est mAP=f(IoU)
}
Let $\{x,y\}$ the coordinates of a ground truth bounding box corner, $\{x_1,y_1\}$ the corresponding \trainmdl
prediction, and $\{x_2=x_1\pm\varepsilon_{\iou},y_2=y_1\pm\varepsilon_{\iou}\}$ the corresponding~\implmdl prediction, we can bound $\iou^{(2)}\ge\iou^{(2)}_{min}$~
\footnote{Detailed equations are in Appendix~\ref{sec:objdet}.}:
$$ \iou^{(2)}_{min} =
\begin{cases}
  \frac{(x_1 - \varepsilon_{\iou})(y_1 - \varepsilon_{\iou})}{xy} & \text{if } x_1 \le x, y_1 \le y \\
  \frac{xy}{(x_1 + \varepsilon_{\iou})(y_1 + \varepsilon_{\iou})} & \text{if } x_1 \ge x, y_1\ge y \\
  \frac{1}{\frac{x_1 + \varepsilon_{\iou}}{x}+\frac{y}{y_1 - \varepsilon_{\iou}}-1} & \text{if } x_1 \ge x, y_1 \le y \\
  \frac{1}{\frac{y_1 + \varepsilon_{\iou}}{y}+\frac{x}{x_1 - \varepsilon_{\iou}}-1} & \text{if } x_1 \le x, y_1 \ge  y
\end{cases}
$$
$$
\iou^{(2)} = \iou^{(2)}_{min} \text{when}  \\
\begin{cases}
x_2 = x_1 + sign(x_1-x)\varepsilon_{\iou}\\
y_2 = y_1 + sign(y_1-y)\varepsilon_{\iou}
\end{cases}
$$



The $\mAP_{\iou \ge t}$ metric considers $\mAP$ when $\iou\ge t~\%$. 
\begin{align*}
&\mAP_{\iou \ge t}(f,\hat f_2) \ge R_{\mAP} \\
&g_{\mAP} = \mAP_{\iou \ge t}(f,\hat f_1) - \mAP_{\iou^{(2)}_{min}\ge t}(f,\hat f_2) \\
&\mAP(f,\hat f_1) \ge R_{\mAP} + g_{\mAP}
\end{align*}

To find the greater $\varepsilon$ for a desired minimum $\mAP$,
one can, similarly to classification tasks,
compute $\mAP_{\iou^{(2)}_{min}\ge t}$ for a set of $\varepsilon_{\iou}$ in a LUT and find $\varepsilon_{\iou} = interp(LUT(\mAP))$.




\subsection{ONNX to support MLMD}
\label{sec:onnx}
\onnx is an open source framework specialized in the inference phase.
It relies on three components:
\onnx core to describe an ML model,
\onnx Runtime~(ORT) to execute \onnx models,
and \onnx script to transform ML model descriptions.
Let us briefly review the \onnx core.

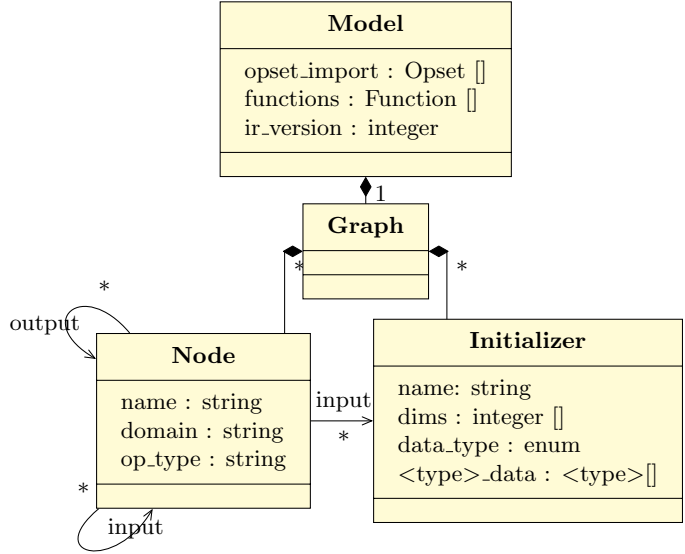
\begin{figure}[hbt]
    \centering
	\resizebox{0.5\linewidth}{!}{
		\begin{tikzpicture}
		\umlclass[y=2,x=0]{Model}{
		  opset\_import : Opset [] \\
		  functions : Function [] \\
		  ir\_version : integer
		}{}
		\umlclass[y=-0.2,x=0]{Graph}{
		}{}
		\umlclass[y=-2.5,x=-2.2]{Node}{
		  name : string \\
		  domain : string \\
		  op\_type : string
		}{}
		\umlclass[y=-2.5,x=2.2]{Initializer}{
		  name: string \\
		  dims : integer [] \\
		  data\_type : enum \\
		  \textless type\textgreater \_data  : \textless type\textgreater []
		}{}
		\umlunicompo[geometry=|-, mult1=1, pos1=-1, mult2=1, pos2=0, align2=left]{Model}{Graph}
		\umlunicompo[geometry=-|-, mult1=*, pos1=1, mult2=*, pos2=0, align2=left]{Graph}{Node}
		\umlunicompo[geometry=-|-, mult1=*, pos1=1, mult2=*, pos2=2.9, align2=left]{Graph}{Initializer}
		\umluniassoc[geometry=-|-, arg1=input, mult1=*, pos1=0.6, mult2=1, pos2=3, align2=left]{Node}{Initializer}
		\umluniassoc[arg=input, mult=*, pos=0, angle1=220, angle2=240, loopsize=1cm]{Node}{Node}
		\umluniassoc[arg=output, mult=*, pos=0.3, angle1=130, angle2=150, loopsize=1cm]{Node}{Node}

		\end{tikzpicture}
	}
    \caption{ONNX simplified UML model}
    \label{fig:onnxgraph}
\end{figure}

\onnx~\cite{ONNX}
 standardizes in particular linear algebra operators and an Intermediate Representation (IR) to specify an ML model structure.
The schematic and simplified representation of the \onnx IR
 is shown in the UML model of Figure \ref{fig:onnxgraph}.
A Model comes with its associated Graph which contains a topologically-ordered set of Nodes, and a set of Initializer values.
A Node defines an $Operator_f,_v$ ($op\_type$), a $domain$, a unique $name$ identifier,
and some attributes to particularize its $Operator_f,_v$.
$Operator_f,_v$ uniquely identifies a \emph{function} $f$ 
with its version $v$.

The Graph structure can be constructed through the Node edges which symbolically reference Tensors.
Two Nodes are connected if they reference the same Tensor name, the source as an output and the target as an input.
The \onnx semantics states that a Node processes its \emph{input} Tensors
and produces \emph{output} Tensors according to its $Operator_f,_v$ semantics.

Initializer values represent constant data (trained parameters : weights and biases), and are referenced by Node inputs.
Those initializer Tensors are identified by a unique name, an array of dimensions ($dims$), a $data\_type$.
A Model also refers to Operator sets ($opset\_import$) which identify a domain name
(e.g. \emph{ai.onnx} is sufficient for neural networks)
 and a set of $Operator_{f,v}$.
A Model also identifies its \onnx file encoding format version ($ir\_version$).
        
\onnx specifies a mathematical semantics at SL0, compliant
more or less with Objective~\ref{obj:semantic}.
However, the definition of core $Operator_{f,v}$ functions is sometimes lacking critical details.
The format also falls short of SL1:
while it captures the data types of Tensors and Nodes, it fails to define the expected behaviour of operations outside the range of the selected representations.
This is considered outside the scope of this paper.
A working group SONNX \footnote{\url{https://github.com/ericjenn/working-groups/tree/ericjenn-srpwg-wg1/safety-related-profile}}
has initiated
the definition of a safety related profile which will refine operators' formal semantics.

\subsection{MLMD \emph{Build} methods}
\label{sec:method}

\begin{figure}[hbt]
	\centering
	\includegraphics[scale=0.3]{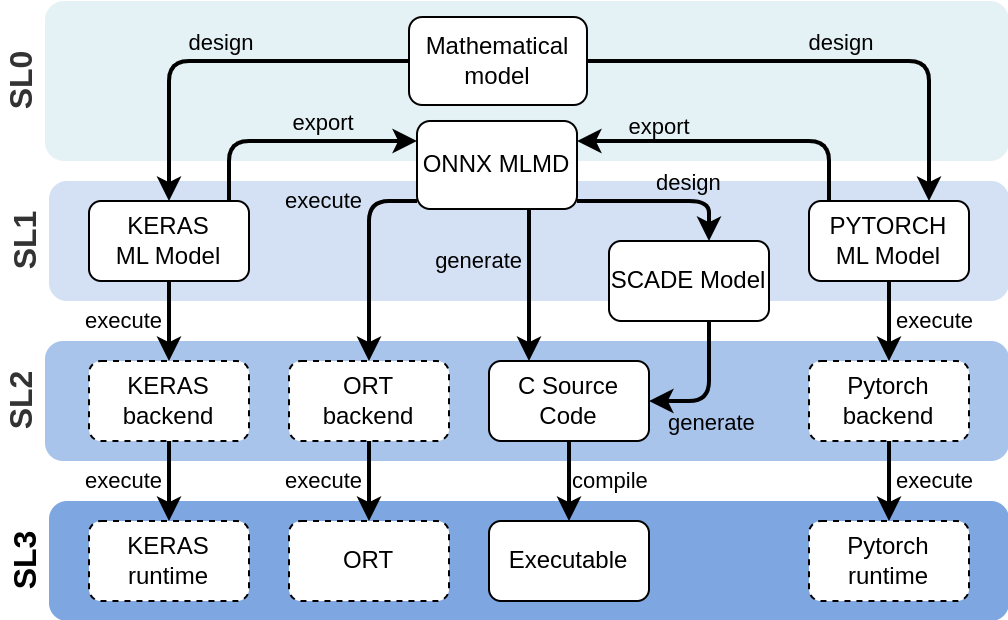}
	\caption{MLMD \emph{Build} and Implementation\label{fig:export}}
\end{figure}

We illustrate in
Figure~\ref{fig:export}
the different layers involved in deriving a~\implmdl.
The graph vertices are ML model representation and environments, and the edges are SL transformations.
On the top level,
the designer selects an ML model (e.g.\ YoLo) that they know to work well for the
task they consider (e.g. object detection).
Such models are defined in the literature
using mathematical operators (SL0).

Then, the applicant selects their favourite training framework,
e.g.\ Keras or PyTorch (respectively on the left and right hand side of the Figure)
or they may even reuse a pre-trained model.
The training associated activities occur within the training framework,
which should capture sufficient SL2+ properties to execute the model.
Accounting for the training environment (framework, backend, runtime, etc.), a \trainmdl is semantically defined at SL3.

The MLMD is then constructed from the \trainmdl. As stated in the introduction, the applicant
can fully reverse engineer the training framework to try and replicate the SL3 \trainmdl\@, or try and replicate the
exact training environment in the~\implmdl.
However, this training environment is hardy a good fit for deployment into resource-constrained, safety-critical environment.
ONNX models can also be executed using ORT,
which would lead to another SL3 branch.
But what we propose instead is to export an ONNX MLMD, and to use code generators from the ONNX description.

Due to the size of ML models, manually deriving the \onnx description or its implementation would be both error-prone and time-consuming.
Thus, we rely on tools available in the \trainmdl and \onnx eco-system to build such MLMD from various training frameworks.
To build an \onnx model, any exporter tool performs design choices to select \onnx operators, graph, parameters~(initializers) matching the semantics of the \trainmdl.
Exporters might omit higher SLx properties in the process.
Code generators similarly take as an input a model and output a (higher) SL model in the process.
They generate code to perform each $Operator_{f,v}$, and the implement the overall data and control flow of the \onnx graph in the generated language and environment.
Without a qualified transformer, one with sufficient experience in service, or a review of the original and transformed model, it is unclear that the transformed model is an equivalent representation or a refinement of the original at a higher SL\@.

\section{Experimental setup}
\label{sec:setup}

To evaluate the semantic preservation of ML models~(\S~\ref{sec:arp-preservation}),
we consider industrial use cases and their deployment as~\implmdl.
We further automate MLMD \emph{Build} and \emph{Replication} methods using tools available in the literature to
assess how to derive and verify SL3~\implmdl, from a \trainmdl\@.
This results in a number of experimental configurations, each denoted as follows:
\begin{center}
    \usecase{mdl}{exp}{repr}{gen}{env}
\end{center}
The notation identifies the considered use case, and the \emph{Build} method, i.e.\ technical choices from the \trainmdl~(SL1) to the~\implmdl~(SL3) as highlighted in Figure~\ref{fig:export}:
\begin{itemize}
    \item \ucf{mdl} identifies the considered use case and \trainmdl\@.
    \item \ucf{exp} identifies the exporter from the \trainmdl to \onnx (SL1).
    \item \ucf{repr} identifies the machine representation (SL1).
    \item \ucf{gen} identifies the code generator (SL2).
    \item \ucf{env} identifies the execution environment, that is the compiler, its configuration, and the hardware
    platform~(SL3).
\end{itemize}
We introduce the considered configurations for each step of the \emph{Build} in the following.
Note that a \texttt{*} as configuration values may be used to denote a set of configuration, e.g. \usecase{mdl}{*}{*}{*}{x86} matches all model \ucf{mdl} configurations on the \ucf{x86}.

\subsection{ML models (\ucf{mdl})}

We present two use cases, intended to perform a regression task for helicopter avionics.
The \trainmdl for use cases are specified in the \keras training framework.
Both use the \ucf{FP32} machine number representation.

\subsubsection{\uclstm}
The \uclstm~ model is a Recurrent Neural Network (Figure~\ref{fig:uc-lstm}), used to implement an aircraft weight estimator~\cite{weight}.
It is composed of 3 bidirectional Long Term Short Term Memory  (LSTM) operators followed by a dense layer.
It takes a ~20 input feature vector and has 1 scalar output.
The LSTM time frame is made of 16 samples.

\subsubsection{\uclinear}
\begin{figure}
\centering
\begin{subfigure}{0.45\linewidth}
    \begin{tikzpicture}[thick,scale=0.5]
        \tiny
        \tikzstyle{block} = [draw, inner sep=3pt];
        \tikzstyle{txt} = [text centered, inner sep=0pt];
        \path (0,0) node[block,rotate=90] (lstm1) {$LSTM_1$};
        \path (lstm1)+(1.5,0) node[block,rotate=90] (lstm2) {$LSTM_2$};
        \path (lstm2)+(1.5,0) node[block,rotate=90] (lstm3) {$LSTM_3$};
        \path (lstm3)+(1.5,0) node[block,rotate=90] (dense4) {$Dense_4$};

        \draw[->] (-1,0) --   (lstm1);
        \draw[->] (lstm1) --  (lstm2);
        \draw[->] (lstm2) --  (lstm3);
        \draw[->] (lstm3) -- (dense4);
        \draw[->] (dense4) -- ($(dense4)+(1,0)$);
    \end{tikzpicture}
    \caption{\ucf{lstm} use case}
    \label{fig:uc-lstm}
\end{subfigure}
\hfill
\begin{subfigure}{0.45\linewidth}
    \begin{tikzpicture}[thick,scale=0.5]
        \tiny
        \tikzstyle{block} = [draw, inner sep=3pt];
        \tikzstyle{txt} = [text centered, inner sep=0pt];
        \path (0,0) node[block,rotate=90] (lstm1) {$Dense_1$};
        \path (lstm1)+(1.5,0) node[block,rotate=90] (lstm2) {$ReLU_2$};
        \path (lstm2)+(1.5,0) node[block,rotate=90] (lstm3) {$Dense_3$};
        \path (lstm3)+(1.5,0) node[block,rotate=90] (dense1) {$ReLU_4$};
        \path (dense1)+(1.5,0) node[block,rotate=90] (dense2) {$Dense_5$};

        \draw[->] (-1,0) --   (lstm1);
        \draw[->] (lstm1) --  (lstm2);
        \draw[->] (lstm2) --  (lstm3);
        \draw[->] (lstm3) -- (dense1);
        \draw[->] (dense1) -- (dense2);
        \draw[->] (dense2) -- ($(dense2)+(1,0)$);
    \end{tikzpicture}
    \caption{\ucf{linear} use case}
    \label{fig:uc-linear}
\end{subfigure}

    \caption{Use cases ML model architecture}
    \label{fig:uc}
\end{figure}
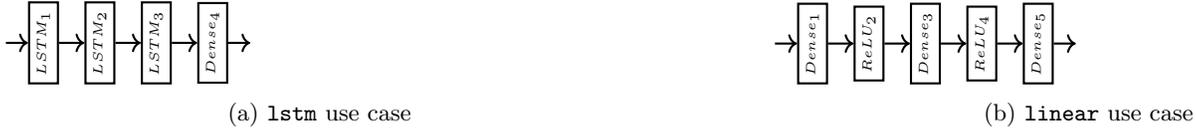

The \uclinear~ use case was introduced in~\cite{del_cistia_gallimard_direct_2023} to compute aircraft loads.
Its structure, depicted in Figure~\ref{fig:uc-linear}, is a Multi Layer Perceptron (MLP) composed of 3 Dense operators, and ReLU activation function.


\subsection{\onnx MLMD (\ucf{exp})}
\label{sec:MLMD_implantation}

Once the \trainmdl has been trained, the MLMD is \emph{built} using the \ucf{legacy} \keras to \onnx exporter.
We then apply automated or manual transformations to the exported model using \onnx script.
All transformations are SL1-preserving.
Indeed, the exported \onnx model cannot be used as is with legacy code generators, which do not support some of the exported Nodes $Operator_{f,v}$ or their parameters.
It is allways necessary to tweak the MLMD to be compliant to the subset of the code generator supported \onnx operators.

\subsection{Code generators (\ucf{gen} and \ucf{repr})}
\label{subsec:code-generators}


We consider the following open source code generators
 to generate C code in our evaluation:
\ucf{onnx2c}~\footnote{\url{https://github.com/kraiskil/onnx2c}}, et
\ucf{acetone}~\cite{silva_acetone_2022}.
Neither code generator,
nor any of the supplemental generators we considered
support the LSTM \onnx operator used in the \uclstm~model.
To replicate the \uclstm~use case, we manually designed a \emph{ANSYS Scade 6}~\cite{scade} model from the \onnx model, and used the \ucf{scade} C code generator.
The \ucf{onnx2c}, \ucf{acetone} and \ucf{scade} code generator carry over the \ucf{FP32} numerical machine representation used in the \trainmdl and in the exported \onnx.

We implemented a mixed precision \onnx runtime which is able to perform inference
using various machine representations to assess their impact on semantic preservation.
The \ucf{ort} generator supports the $orep =$\{\ucf{FP64}, \ucf{FP32}, \ucf{FP16}, \ucf{BF16}, \ucf{INT16}, \ucf{INT14}, \ucf{INT12}, \ucf{INT10}\} data types;
\ucf{FP}$b$ are defined by \ieee,
\ucf{BF16} is the truncated \ucf{FP32},
\ucf{INT}$b$ are quantized integer in $\mathbb{Z}_b$.
The floating point formats \ucf{FP64}, \ucf{FP32}, \ucf{FP16}, \ucf{BF16} respectively offer 52, 24, 11, 8 bits of mantissa precision.

We thus have three distinct sets of \ucf{gen} configurations:
\begin{itemize}
    \item \usecase{*}{legacy}{R}{ort}{*}: either use-case running on the modified ORT
    (with \ucf{R}~$\in orep$);
    \item \usecase{linear}{legacy}{FP32}{C}{*}: \ucf{linear} state-of-the-art code generators
     (with \ucf{C}~$\in \{$\ucf{onnx2c}, \ucf{acetone}$\}$);
    \item \usecase{lstm}{legacy}{FP32}{scade}{*}: \ucf{lstm} using the \ucf{scade} code generator.
\end{itemize}

\subsection{Execution environment (\ucf{env})}



We consider two execution environments for our evaluation,
a Linux-based Intel server~(\ucf{x86}),
and an NXP T1042 PowerPC~(\ucf{t1042}).
The \ucf{t1042} is designed for embedded systems
and it does not support the \ort runtime, only configurations that went through a C code generator (\ucf{scade}, \ucf{onnx2c}, or \ucf{acetone} with per use-case restrictions).
The generated C code is compiled
using conservative compiler options (no optimization).
The Windriver\textregistered\ DIAB compiler is used on the \ucf{t1042} platform.

We thus explore a subset of the following configurations:
\begin{itemize}
    \item \usecase{*}{legacy}{FP32}{*}{t1042}: either use-case running generated C-code on the \ucf{t1042} platform;
    \item \usecase{*}{legacy}{*}{ort}{x86}: either use-case running ORT on the \ucf{x86} platform.
\end{itemize}


\section{Results}
\label{sec:results}


The evaluation of the semantics preservation for the considered configurations relies on the approach proposed in Section~\ref{sec:applicability}.
We first consider for each use-case the properties satisfied by the \trainmdl,
and the corresponding budget $g_M$ for various metrics $M$, 
to compute the acceptable error $\varepsilon_M$ between the \trainmdl and the~\implmdl.
We then assess for each experimental configuration whether it fits said requirements, 
i.e. whether the~\implmdl satisfies the same properties as the \trainmdl\@.

\subsection{Use case \uclstm}
\subsubsection{\trainmdl verification}

\begin{figure}[!h]
		   \centering
		   \begin{subfigure}{0.49\linewidth}
                     \centering
			   \includegraphics[width=.65\linewidth]{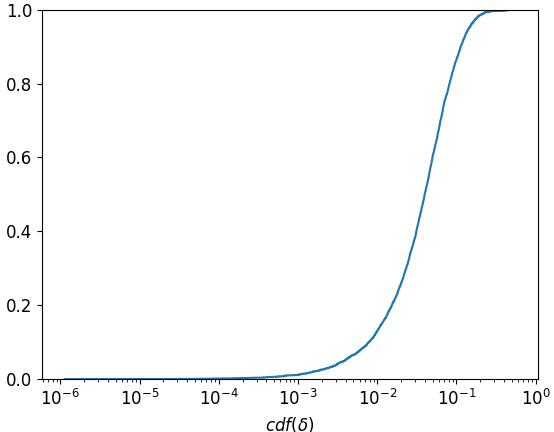}
			   \caption{\uclstm\ use case}
			   \label{fig:delta_cdf-lstm}
		   \end{subfigure}
		   \begin{subfigure}{0.49\linewidth}
                     \centering
			   \includegraphics[width=.65\linewidth]{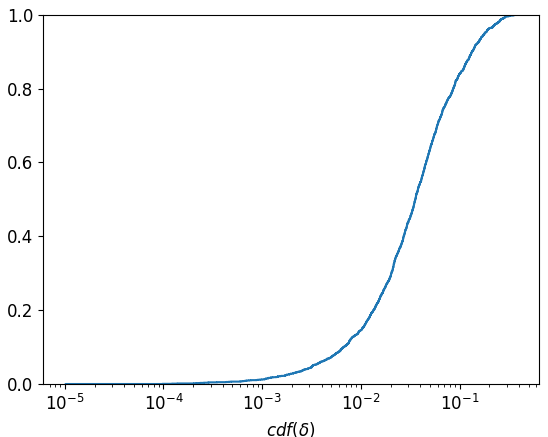}
			   \caption{\uclinear\ use case}
			   \label{fig:delta_cdf-linear}
		   \end{subfigure}
		   \caption{Cumulative distribution of errors between the \trainmdl and ground truth ($cdf(\hat f_1(x_i) - f(x_i))$)}
		   \label{fig:delta_cdf}
\end{figure}

The verification dataset is composed of 4288 samples.
We first compute the distribution of errors between the \trainmdl and the ground truth, $\hat f_1(x_i) - f(x_i)$.
Figure~\ref{fig:delta_cdf-lstm} presents the corresponding cumulative distribution function of errors.
For each error value $x$, the function provides the portion of the dataset (on the y-axis) with an error less than or equal to $x$.

The regression metrics for \ucf{lstm} are computed in Table~\ref{tab:lstm-metrics}.
For each metric $M$, the table captures the performance of the \trainmdl ($M^{(1)}$), the acceptable value ($R_M$)
provided by the model designer, and a bound on the error between the \trainmdl and the~\implmdl predictions ($\varepsilon_M$) to ensure semantic preservation.
Metric $R^2$ provides the most stringent requirement,
$\varepsilon_{M} = 0.008$.

\begin{table}[hbt]
	\centering
	\small
	\begin{tabular} { |l|r|l|l| }
		\hline
		$M$  &
		$M^{(1)}$  & 
		$R_M$ &
		$\varepsilon_{M}$ \\
		\hline
		$MAX$  & $0.55$ & $\leq 1.0$ &  $\leq 0.44$ \\
		$\mae$  & $0.053$ & $\leq 0.07$ &  $\leq 0.017$ \\
		$\mse$  & $0.005$ & $\leq 0.006$  & $\leq 0.014$\\
		$\mape$ & $0.079$ & $\leq 0.09$  & $\leq 0.009$\\
		$\rsq$  & $0.841$ & $\geq 0.83$ & $\leq \textbf{0.008}$ \\
		$\evs$  & $0.849$ & $\geq 0.83$  & $\leq $ 0.013\\
		$\var$  & $0.005$ & $\leq 0.01$  & $\leq 0.03$ \\
		$\bias$ & $-0.017$ & $\leq 0.03$  & $\leq 0.012$ \\
		\hline
	\end{tabular}
	\captionof{table}{Regression metrics for \uclstm}\label{tab:uclstm-regression}
	\label{tab:lstm-metrics}
\end{table}

\subsubsection{\implmdl verification}

\begin{figure}
    \centering
    \includegraphics[scale=0.55]{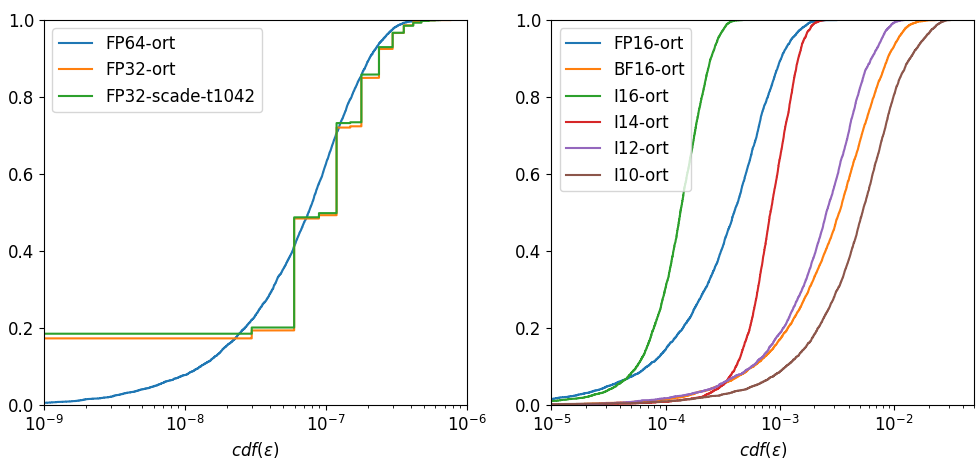}
	\caption{Cumulative distribution function of errors between the \trainmdl and~\implmdl
		$cdf(\hat f_2(x_i) - \hat f_1(x_i))$}
	\label{fig:cdf}
\end{figure}

\todo{BL: Pas clair sur quel dataset on compare le \trainmdl et \implmdl.}
\todo{BL: Pourquoi on a deux FP32 dans la Figure~\ref{fig:cdf} à gauche?}

We now consider the \implmdl for the various experimental configurations in comparison to the \trainmdl\@.
We first present the cumulative distribution function of the error $\varepsilon_i = \hat f_2(x_i) - \hat f_1(x_i)$
between the \trainmdl and each \implmdl in Figure~\ref{fig:cdf}.
Each curve represents a different configuration identified by its type (\ucf{ort} on \ucf{x86}) or platform (
\ucf{t1042}).
Configurations using a reduced numerical precision and thus increased error, \ucf{INT}$_b$ or 16-bit floating points, are plotted separately on the right of Figure~\ref{fig:cdf}.

We see that the errors of the C generated source codes running on the \ucf{t1042} environment are very close to that of our \onnx runtime implementation which hints at a similarity between semantics.
Unsurprisingly, the error increases (from left to right) as the data type loses precision.

The final assessment is to check for each candidate configuration
\usecase{lstm}{legacy}{*}{*}{*}
if it is in the margins of the most constraining $R_M$ of Table~\ref{tab:uclstm-regression}, i.e.
  $|\varepsilon_i| < \varepsilon_{M} \forall i \in [1,n]$.
The results are presented in Table~\ref{tab:metrics_lstm_eps} capturing for each configuration the maximum observed error ($max(|\varepsilon_i|)$), and its comparison to $\varepsilon_{M}$.
Only the most reduced numerical precisions \ucf{BF16}, \ucf{INT12}, and \ucf{INT10} fail the assertion.
All other candidates (in particular the one actually running on the \ucf{t1042} environment) replicate the semantics of the \trainmdl\@.

\begin{table}[hbt]
	\centering
	 \small
	\begin{tabular} { lll }
	 \toprule
	\implmdl & $\max(|\varepsilon_i|)$ & $\leq \varepsilon_{M}$  \\
	 \hline
\usecase{*}{*}{FP64}{ort}{x86} & $6e^{-7}$ & Y \\
\usecase{*}{*}{FP32}{ort}{x86} &$7e^{-7}$ & Y  \\
\usecase{*}{*}{FP32}{scade}{t1042} & $7e^{-7}$ & Y  \\
\usecase{*}{*}{FP16}{ort}{x86}& $3e^{-3}$ & Y  \\
\usecase{*}{*}{BF16}{ort}{x86}& $2e^{-2}$ & N  \\
	 \hline
\usecase{*}{*}{INT16}{ort}{x86}& $4e^{-4}$ & Y  \\
\usecase{*}{*}{INT14}{ort}{x86}& $2e^{-3}$ & Y  \\
\usecase{*}{*}{INT12}{ort}{x86}& $2e^{-2}$ & N  \\
\usecase{*}{*}{INT10}{ort}{x86}& $3e^{-2}$ & N  \\
 \bottomrule
 	\end{tabular}
	\captionof{table}{Replication metrics for \ucf{lstm} with $\varepsilon_{M} = 0.008$}\label{tab:metrics_lstm_eps}
\end{table}

\subsection{Use case \uclinear}
\subsubsection{\trainmdl verification}

The verification dataset is composed of 2000 samples.
We perform a similar assessment for use-case \ucf{linear}.
Figure~\ref{fig:delta_cdf-linear} presents the cumulative distribution function of errors between the \trainmdl and the ground truth.
The regression metrics and $\varepsilon_M$ bounds on the \trainmdl and \implmdl prediction errors for~\ucf{linear} are
presented in Table~\ref{tab:metrics_linear_eps}.
The most constraining $R_M$ are $EVS$ and once again $R^2$ with an $\varepsilon_{M} = 0.015$.

\begin{table}[hbt]
	\centering
	\small
	\begin{tabular} { |l|r|l|l| }
		\hline
		$M$  &
		$M^{(1)}$  & 
		$R_M$ &
		$\varepsilon_{M}$ \\
		\hline
		$\mae$  & $0.033$ & $\leq 0.06$ &  $\leq 0.026$ \\
		$\mse$  & $0.002$ & $\leq 0.01$  & $\leq 0.088$\\
		$\rsq$  & $0.821$ & $\geq 0.8$ & $\leq \textbf{0.015}$ \\
		$\evs$  & $0.821$ & $\geq 0.8$  & $\leq \textbf{0.015}$\\
		$\var$  & $0.002$ & $\leq 0.01$  & $\leq 0.088$ \\
		$\bias$ & $-0.0003$ & $\leq 0.03$  & $\leq 0.029$ \\
		\hline
	\end{tabular}
	\captionof{table}{Regression metrics for \uclinear}\label{tab:ucllinear-regression}
\end{table}

\subsubsection{\implmdl verification}
Table~\ref{tab:metrics_linear_eps} presents the assessment for each configuration of whether it satisfies to the semantic preservation of the \ucf{linear} \trainmdl\@.
Taking into account the tightest constraint in Table~\ref{tab:ucllinear-regression},
both \ucf{t1042}~\implmdl  are also compliant.
Similarly, the higher precision representations \ucf{FP32}, \ucf{FP16}, and \ucf{INT16} \ucf{ort}~\implmdl are
compliant.

\begin{table}[hbt]
	\centering
	 \small
	\begin{tabular} { lll }
	 \toprule
	\implmdl & $\max(|\varepsilon_i|)$ & $\leq \varepsilon_{M}$\\
	 \hline 	
\usecase{*}{*}{FP32}{ort}{x86} & $3e^{-7}$ & Y  \\
\usecase{*}{*}{FP32}{onnx2c}{t1042} & $7e^{-7}$ & Y \\
\usecase{*}{*}{FP32}{acetone}{t1042} & $7e^{-7}$ & Y \\
\usecase{*}{*}{FP16}{ort}{x86} & $2e^{-3}$ & Y  \\
\usecase{*}{*}{BF16}{ort}{x86} & $1e^{-2}$ & N \\
 \hline
\usecase{*}{*}{INT16}{ort}{x86} & $2e^{-3}$ & Y \\
\usecase{*}{*}{INT14}{ort}{x86} & $1e^{-2}$ & N \\
\usecase{*}{*}{INT12}{ort}{x86} & $3e^{-2}$ & N \\
\usecase{*}{*}{INT10}{ort}{x86} & $1e^{-1}$ & N \\
 \bottomrule
 	\end{tabular}
	\captionof{table}{Replication metrics for \ucf{linear} with $\varepsilon_{M} = 0.015$}\label{tab:metrics_linear_eps}
\end{table}

\section{Related work}
\label{sec:related-work}

Model exporter tools are designed to translate the training framework semantics into the MLMD semantics.
However, varied or repeated translations steps 
can lead to numerous compatibility and replication issues, reported in the survey~\cite{jajal2024interoperability}.
\cite{Geyer2024EfficientAM} further highlighted the tradeoff between model inference performance and precision when using hardware accelerators.

In the ML community,
the \onnx exchange format is
one of the most widely-supported by popular training frameworks.
Prior works have built MLMD with NNEF~\cite{FormalDescMLmodel}
and \onnx (Open Neural Network eXchange)~\cite{ACASXu}.
However, these approaches did not explore 
how to export a TFM, and how to verify its compliance to \arpshort objectives.
\cite{ACASXu} did consider bit-accurate replication which proved costly considering the number of moving parts involved in the execution of the TFM.

Numerous frameworks can generate C, C++ or CUDA code compatible with airborne systems development life-cycle out of a MLMD~\cite{silva_acetone_2022, Cerioli2024NeuralCastingAF,perotto}.
Indeed, in line with legacy guidance \doseventi, producing source code from the MLMD is a certification-friendly approach.
In this work, we rely on implementations produced by C/C++
code generators (\ucf{acetone}, \ucf{onnx2c}, \ucf{scade}) for selected target environment (\ucf{x86}, \ucf{t1042}).
We have chosen code generator methods which are widely adopted in the field of safety critical avionics.
The target environment in any case will generate different results than the TFM due to computation roundings studied in~\cite{Fang2024StatisticalRE} and~\cite{beuzeville}, and activation functions algorithm approximations.



\section{Conclusion}
\label{sec:conclusion}



In legacy avionics development, industrialization and certification require that verification of requirements or properties is performed in the~\implmdl,
unless an argument is provided to justify that verification which were performed on a different execution environment are still valid in the~\implmdl.
One argument could be a bit-accurate replication between \trainmdl and the~\implmdl which is a quite hard constraint on the development. 
To avoid this constraint, we propose a ML model semantics preservation definition and a supporting verification methodology, based on state-of-the-art ML metrics.
This method relies on two independent processes:
\begin{enumerate*} 
\item the verification of the~\trainmdl metrics \emph{including budget $g_M$}, and using a test set including the ground truth,
\item  the verification that the predictions of the~\implmdl fit into \emph{error margin} $\varepsilon_M$ using a chosen set of \trainmdl predictions. This second process does not require any knowledge of the metrics, nor the ground truth. It takes as input the MLMD including $\varepsilon_{M}$. 
Its objective is to demonstrate an upper bound of the cumulated rounding and approximations errors of the algorithm.
In this paper we used the test set input to sample the \trainmdl for the sake of simplicity, but future research might elaborate a sound proof of $\varepsilon_{M}$ threshold. The probability to exceed $\varepsilon_M$ threshold can be bounded: $P(\varepsilon > \varepsilon_M) < P_M$ where $P_M$ is a threshold included in MLMD with $\varepsilon_M$. This might require a specific input space sampling. The Objective IMP-08 of  the concept paper \cite{Easaconcept} is limiting verification to the test set which might not be sufficient nor appropriate for implementation verification.
\end{enumerate*}

We further decomposed the level of details captured by ML model representations into semantic levels, related to these margins.
We considered how existing tools~\cite{silva_acetone_2022, ikarashi_exocompilation_2022} could support the definition of (semi-)automated methods to \emph{Build} a \implmdl and replicate a \trainmdl,
and we showed that they can indeed preserve the properties of industrial ML use cases~\cite{del_cistia_gallimard_direct_2023, caroline_del_cistia_gallimard_direct_nodate, jouve_estimation_2023} from the \trainmdl into a~\implmdl.





\section*{Acknowledgements}
This work has benefitted from the AI Interdisciplinary Institute ANITI. ANITI is funded by
the France 2030 program under the Grant agreement n°ANR-23-IACL-0002.

\bibliographystyle{abbrv}
\bibliography{main}

\begin{thebibliography}{10}

\bibitem{IEEE754}
{IEEE} {Standard} for {Floating}-{Point} {Arithmetic}.
\newblock {\em {IEEE Std 754-2008}}, pages 1--70, 2008.

\bibitem{dragoonBook}
A.~V. Aho, M.~S. Lam, R.~Sethi, and J.~D. Ullman.
\newblock {\em Compilers: Principles, Techniques, and Tools (2nd Edition)}.
\newblock {Addison Wesley}, August 2006.

\bibitem{ONNX}
J.~Bai, F.~Lu, K.~Zhang, et~al.
\newblock {ONNX: Open Neural Network Exchange}.
\newblock \url{https://onnx.ai/}, 2019.

\bibitem{beuzeville}
T.~Beuzeville.
\newblock {\em {Backward error analysis of artificial neural networks with
  applications to floating-point computations and adversarial attacks}}.
\newblock PhD thesis, {Universit{\'e} de Toulouse}, 2024.

\bibitem{Cerioli2024NeuralCastingAF}
A.~Cerioli, R.~Miccini, C.~Laroche, T.~Piechowiak, L.~Pezzarossa, J.~Spars{\o},
  and M.~Schoeberl.
\newblock {NeuralCasting: A Front-End Compilation Infrastructure for Neural
  Networks}.
\newblock {\em 2024 11th International Conference on Internet of Things:
  Systems, Management and Security (IOTSMS)}, 2024.

\bibitem{Christensen2024TowardsCA}
J.~M. Christensen, W.~Zaeske, J.~Beck, S.~Friedrich, T.~Stefani, A.~A. Girija,
  E.~Hoemann, U.~Durak, F.~K{\"o}ster, T.~Kr{\"u}ger, and S.~Hallerbach.
\newblock {Towards Certifiable AI in Aviation: A Framework for Neural Network
  Assurance Using Advanced Visualization and Safety Nets}.
\newblock {\em 2024 AIAA DATC/IEEE 43rd Digital Avionics Systems Conference
  (DASC)}, pages 1--9, 2024.

\bibitem{caroline_del_cistia_gallimard_direct_nodate}
C.~del Cistia~Gallimard, K.~Nikolajevic, F.~Beroul, J.~Denoulet, B.~Granado,
  and C.~Marsala.
\newblock {Direct Load Recognition to Estimate the Damper Load on the H175
  Fleet}.
\newblock In {\em {European Rotorcraft Forum}}, 2023.

\bibitem{scade}
B.~Dion, M.~Najork, N.~Dalmasso, J.-L. Colaço, O.~Andrieu, B.~Buettner,
  T.~Most, and J.~R. de~Amaral.
\newblock {Programming Safety-Critical Neural Network Inference Models}.
\newblock In {\em {11th European Congress on Embedded Real Time Software and
  Systems (ERTS)}}, 2022.

\bibitem{TowardsDesignAssuranceLevelC}
K.~Dmitriev, J.~Schumann, and F.~Holzapfel.
\newblock {Towards Design Assurance Level C for Machine-Learning Airborne
  Applications}.
\newblock In {\em 2022 IEEE/AIAA 41st Digital Avionics Systems Conference
  (DASC)}, pages 1--6, 2022.

\bibitem{Easaconcept}
{EASA}.
\newblock {Concept Paper: guidance for Level 1 \& 2 machine learning
  applications - Proposed Issue 02}, 2024.

\bibitem{Elboher2024RobustnessAO}
Y.~Y. Elboher, R.~Elsaleh, O.~Isac, M.~Ducoffe, A.~Galametz, G.~Pov{\'e}da,
  R.~Boumazouza, N.~Cohen, and G.~Katz.
\newblock {Robustness Assessment of a Runway Object Classifier for Safe
  Aircraft Taxiing}.
\newblock {\em 2024 AIAA DATC/IEEE 43rd Digital Avionics Systems Conference
  (DASC)}, 2024.

\bibitem{Fang2024StatisticalRE}
Y.~Fang and L.~Chen.
\newblock Statistical rounding error analysis for random matrix computations.
\newblock 2025.

\bibitem{floyd1993assigning}
R.~W. Floyd.
\newblock Assigning meanings to programs.
\newblock {\em Mathematical Aspects of Computer Science}, pages 19--32, 1967.

\bibitem{gabreau:hal-WG}
C.~Gabreau and {et al.}
\newblock {EUROCAE WG114 – SAE G34: a joint standardization initiative to
  support Artificial Intelligence revolution in aeronautics}, 2023.
\newblock Keynote of SafeAI.

\bibitem{ACASXu}
C.~Gabreau, M.-C. Teulieres, E.~Jenn, A.~Lemesle, D.~P. Butucaru, F.~Thiant,
  L.~Fischer, and M.~Turki.
\newblock {A study of an ACAS-Xu exact implementation using ED-324/ARP6983}.
\newblock In {\em {12th European Congress on Embedded Real Time Software and
  Systems (ERTS)}}, 2024.

\bibitem{del_cistia_gallimard_direct_2023}
C.~D.~C. Gallimard, F.~Beroul, J.~Denoulet, K.~Nikolajevic, A.~Pinna,
  B.~Granado, and C.~Marsala.
\newblock {Harmonic Decomposition to Estimate Periodic Signals using Machine
  Learning Algorithms: Application to Helicopter Loads}.
\newblock In {\em {International Joint Conference on Neural Networks (IJCNN)}},
  pages 1--8. {IEEE}, 2022.

\bibitem{FormalDescMLmodel}
A.~Gauffriau, I.~De~Albuquerque~Silva, and C.~Pagetti.
\newblock {Formal description of ML models for unambiguous implementation}.
\newblock In {\em {12th European Congress on Embedded Real Time Software and
  Systems (ERTS)}}, 2024.

\bibitem{Geyer2024EfficientAM}
F.~Geyer, J.~Freitag, T.~Schulz, and S.~Uhrig.
\newblock Efficient and mathematically robust operations for certified neural
  networks inference.
\newblock In {\em 6th Workshop on Accelerated Machine Learning (AccML) at
  HiPEAC}, 2024.

\bibitem{hoare}
C.~A.~R. Hoare.
\newblock An axiomatic basis for computer programming.
\newblock {\em Commun. ACM}, page 576–580, 1969.

\bibitem{ikarashi_exocompilation_2022}
Y.~Ikarashi, G.~L. Bernstein, A.~Reinking, H.~Genc, and J.~Ragan-Kelley.
\newblock Exocompilation for productive programming of hardware accelerators.
\newblock In {\em Proceedings of the 43rd ACM SIGPLAN International Conference
  on Programming Language Design and Implementation}, PLDI 2022, page
  703–718, 2022.

\bibitem{jajal2024interoperability}
P.~Jajal, W.~Jiang, A.~Tewari, E.~Kocinare, J.~Woo, A.~Sarraf, Y.-H. Lu, G.~K.
  Thiruvathukal, and J.~C. Davis.
\newblock {Interoperability in Deep Learning: A User Survey and Failure
  Analysis of ONNX Model Converters}.
\newblock In {\em Proceedings of the 33rd ACM SIGSOFT International Symposium
  on Software Testing and Analysis}, pages 1466--1478, 2024.

\bibitem{jouve_estimation_2023}
J.~Jouve, C.~del Cistia~Gallimard, K.~Nikolajevic, and H.~Morel.
\newblock {Estimation of confidence margins for Direct Load Recognition (DLR)
  using supervised and unsupervised machine learning}.
\newblock In {\em Vertical Flight Society}, 2023.

\bibitem{kaakai2022toward}
F.~Kaakai and {et al.}
\newblock Toward a machine learning development lifecycle for product
  certification and approval in aviation.
\newblock {\em SAE International journal of aerospace}, 15, 2022.

\bibitem{kaakai2023datacentric}
F.~Kaakai and {et al.}
\newblock Data-centric operational design domain characterization for machine
  learning-based aeronautical products, 2023.

\bibitem{weight}
A.~Mechouche, A.~Rocher, and V.~Aubin.
\newblock Method for training at least one artificial intelligence model for
  estimating the mass of an aircraft during flight based on utilisation data.
\newblock US Patent US20240005207A1 App. 18/209,183, 2024, European Patent
  EP4300053B1.

\bibitem{mleap}
{MLEAP Consortium, EASA Research}.
\newblock {Machine Learning Application Approval (MLEAP) Final Report}, May
  2024.

\bibitem{metrics}
R.~Padilla, W.~L. Passos, T.~L.~B. Dias, S.~L. Netto, and E.~A.~B. da~Silva.
\newblock A comparative analysis of object detection metrics with a companion
  open-source toolkit.
\newblock {\em Electronics}, 10(3), 2021.

\bibitem{perotto}
F.~S. Perotto, A.~Fernandes~Pires, J.-L. Farges, Y.~Bouchebaba, M.~Belcaid,
  E.~Bonnafous, C.~Pagetti, F.~Boniol, X.~Pucel, A.~Chan-Hon-Tong, S.~Herbin,
  M.~Cassaro, and S.~Kraiem.
\newblock {Thinking the certification process of embedded ML-based aeronautical
  components using AIDGE, a French open and sovereign AI platform}.
\newblock In {\em {International Conference on Cognitive Aircraft Systems}},
  2024.

\bibitem{DO178}
{RTCA/EUROCAE}.
\newblock D{O}-178{C}/{E}{D}-12{C} - {S}oftware {C}onsiderations in {A}irborne
  {S}ystems and {E}quipment {C}ertification, 2011.

\bibitem{denotational}
D.~A. Schmidt.
\newblock Denotational semantics: a methodology for language development.
  william c, 1986.

\bibitem{silva_acetone_2022}
I.~D.~A. Silva, T.~Carle, A.~Gauffriau, and C.~Pagetti.
\newblock {ACETONE:} predictable programming framework for {ML} applications in
  safety-critical systems.
\newblock In {\em 34th Euromicro Conference on Real-Time Systems, {ECRTS} 2022,
  July 5-8, 2022, Modena, Italy}, 2022.

\bibitem{Wasson2024DeobfuscatingML}
K.~S. Wasson and R.~Voros.
\newblock {Deobfuscating Machine Learning Assurance and Approval}.
\newblock {\em 2024 AIAA DATC/IEEE 43rd Digital Avionics Systems Conference
  (DASC)}, pages 1--10, 2024.

\bibitem{kahan}
K.~William.
\newblock Further remarks on reducing truncation errors.
\newblock {\em CACM}, 1965.

\end{thebibliography}
\section{Appendix}
\label{sec:annex}
\subsection{Regression Tasks}
This appendix details how to compute the ML metrics out of $\varepsilon_M$.
The following equations are defining the~\implmdl metric
as a function of the \trainmdl metric and the $\varepsilon_{M}$.
$\delta_i$ represents the error between \trainmdl prediction $\hat f_1(x_i)$ and the ground truth $f(x_i)$, i.e.
$\delta_i = \hat f_1(x_i) - f(x_i)$.
$\varepsilon_i$ represents the error between the~\implmdl prediction $\hat f_2(x_i)$  and the \trainmdl prediction $\hat f_1(x_i)$, i.e.
$\varepsilon_i = \hat f_2(x_i) - \hat f_1(x_i)$.
The metric $M^{(1)}$ is related to $\hat f_1(x)$ and $M^{(2)}$ is related to $\hat f_2(x)$.

\noindent\textbf{MAE metric}
\begin{align*}
\mae^{(2)} &= \frac{1}{n} \sum^n |\hat f_2(x_i) - f(x_i)| = \frac{1}{n} \sum^n |\varepsilon_i + \delta_i| \\
&\leq \mae^{(1)} + \frac{1}{n} \sum^n |\varepsilon_i| \leq \mae^{(1)} + \varepsilon_{\mae}  \\
g_{\mae} &= \max(\mae^{(2)}) - \mae^{(1)} = \varepsilon_{\mae}
\end{align*}

\noindent\textbf{MSE metric}
\begin{align*}
\mse^{(2)} &= \frac{1}{n} \sum^n(\hat f_2(x_i) - f(x_i))^2 =\frac{1}{n} \sum^n (\varepsilon_i + \delta_i)^2\\
&=  \frac{1}{n} \sum^n \varepsilon_i^2 + \delta_i^2 + 2\varepsilon_i  \delta_i =  \mse^{(1)}+\sum^n \varepsilon_i^2+\frac{2}{n} \vec{\varepsilon}.\vec{\delta} \\
\vec{\varepsilon}.\vec{\delta} &\in \left[- n \varepsilon_{\mse}|\bias^{(1)}|, n \varepsilon_{\mse}|\bias^{(1)}|\right]\\
\mse^{(2)} &\leq \mse^{(1)} + \varepsilon_{\mse}^2 + 2\varepsilon_{\mse}|\bias^{(1)}| \\
g_{\mse} &= \max(\mse^{(2)})-\mse^{(1)} = \varepsilon_{\mse}^2 + 2\varepsilon_{\mse}|\bias^{(1)}|
\end{align*}

\noindent\textbf{Variance metric}
\begin{align*}
\var^{(1)} &= \mse^{(1)} - \bias_{(1)}^2 \\
\var^{(2)} &= \mse^{(2)} - \bias_{(2)}^2 \\
&= \frac{1}{n} \sum^n \varepsilon_i^2 + \delta_i^2 + 2\varepsilon_i  \delta_i - (\overline{\varepsilon} + \bias_{(1)})^2 \\
&= \frac{1}{n} \sum^n \varepsilon_i^2 + \delta_i^2 + 2\varepsilon_i  \delta_i - \overline{\varepsilon}^2 - \bias_{(1)}^2 - 2\overline{\varepsilon}  \bias_{(1)}\\
&= \mse^{(1)} - \bias_{(1)}^2 + \frac{1}{n} \sum^n \varepsilon_i^2 + 2\varepsilon_i  \delta_i - \overline{\varepsilon}^2  - 2 \overline{\varepsilon} \overline{\delta}\\
&= \var^{(1)} + \frac{1}{n} \sum^n \varepsilon_i^2 + 2\varepsilon_i  \delta_i - \overline{\varepsilon}^2 - 2 \overline{\varepsilon} \overline{\delta} \\
\var^{(2)} &\leq \var^{(1)} + \varepsilon_{\var}^2 + 2\varepsilon_{\var}|\bias^{(1)}| \\
g_{\var} &= \max(\var^{(2)})-\var^{(1)} = \varepsilon_{\var}^2 + 2\varepsilon_{\var}|\bias^{(1)}|
\end{align*}

\noindent\textbf{Explained variance score metric}
\begin{align*}
\evs^{(2)} &= 1 - \frac{\var^{(2)}}{\var(f(x))} \\
\evs^{(2)} &= \evs^{(1)} - \frac{ \frac{1}{n} \sum^n \varepsilon_i^2 + 2\varepsilon_i  \delta_i - \overline{\varepsilon}^2 - 2 \overline{\varepsilon} \overline{\delta}}{\var(f(x))} \\
\overline{\varepsilon} &\xrightarrow[n\to \infty]{}0, \\
\evs^{(2)} &\geq  \evs^{(1)} - \frac{\varepsilon_{\evs}^2 + 2\varepsilon_{\evs}|\bias^{(1)}|}{\var(f(x))} \\
g_{\evs} &= \evs^{(1)} - \min(\evs^{(2)}) = \frac{\varepsilon_{\evs}^2 + 2\varepsilon_{\evs}|\bias^{(1)}|}{\var(f(x))}
\end{align*}

\noindent\textbf{Coefficient of determination (R2) metric}
\begin{align*}
\rsq_{(2)} &= 1 - \frac{\sum^n (\hat f_2(x) - f(x))^2}{\sum^n (f(x) - \overline{f(x)})^2} 
= 1 - \frac{ \sum^n (\varepsilon_i + \delta_i)^2}{\sum^n (f(x) - \overline{f(x)})^2} \\
&= \rsq_{(1)} - \frac{ \frac{1}{n}\sum^n (\varepsilon_i^2 + 2\varepsilon_i \delta_i)}{\frac{1}{n}\sum^n (f(x) - \overline{f(x)})^2}
= \rsq_{(1)} - \frac{ \frac{1}{n}\sum^n \varepsilon_i^2 + \frac{2}{n} \vec{\varepsilon}\vec{\delta}}{\var(f(x))}\\
\rsq_{(2)} &\geq \rsq_{(1)} - \frac{\varepsilon_{\rsq}^2 + 2\varepsilon_{\rsq}|\bias^{(1)}|}{\var(f(x))} \\
g_{\rsq} &= \rsq_{(1)} - \min(\rsq_{(2)}) = \frac{\varepsilon_{\rsq}^2 + 2\varepsilon_{\rsq}|\bias^{(1)}|}{\var(f(x))}
\end{align*}

\noindent\textbf{MAPE metric}
The MAPE metric is relative to the sample output value. In this case, we redefine $\varepsilon$ and $\delta$ as follows:
$\hat f_1(x_i) = (1+\delta_i) f(x_i)$, similarly: $\delta_i = \frac{\hat f1(x_i) - f(x_i)}{f(x_i)}$.
$\hat f_2(x_i) = (1+\varepsilon_i)\hat f_1(x_i)$, similarly: $\varepsilon_i = \frac{\hat f_2(x_i) - \hat f_1(x_i)}{\hat f_1(x_i)}$.
$\hat f_2(x_i) = (1+\varepsilon_i)(1+\delta_i)f(x_i)$.
Let $(1+\gamma_i) = (1+\varepsilon_i)(1+\delta_i)$,
then
$\hat f_2(x_i) = (1+\gamma_i)f(x_i)$, similarly: $\gamma_i = \frac{\hat f_2(x_i) - f(x_i)}{f(x_i)}$.

\begin{align*}
\mape^{(2)} &=\frac{1}{n} \sum^n\left|\frac{\hat f_2(x_i) - f(x_i)}{f(x_i)}\right| = \frac{1}{n} \sum^n |\gamma_i|\\
&= \frac{1}{n} \sum^n |(1+\varepsilon_i)(1+\delta_i) - 1|\\
&= \frac{1}{n} \sum^n |\varepsilon_i \delta_i + \varepsilon_i+\delta_i|\\
\mape^{2} & \leq  \mape^{(1)} + \frac{1}{n} \sum^n |\varepsilon_i(1+\delta_i)|\\
&\leq  \mape^{(1)} + \varepsilon_{\mape}\frac{1}{n} \sum^n |1+\delta_i|
\end{align*}
\begin{align*}
\mape^{(2)} &\leq \mape^{(1)} +  \varepsilon_{\mape} (1 + \mape^{(1)}) \\
g_{\mape} &=  \max(\mape^{(2)}) - \mape^{(1)} = \varepsilon_{\mape} (1 + \mape^{(1)}) 
\label{eq:mae}
\end{align*}

\subsection{Object detection task}
\label{sec:objdet}

\noindent\textbf{1-dim IoU}
Let us define $x$ the ground truth, $x_1$ the \trainmdl prediction and $x_2$ the~\implmdl predictions in 1-dim.
We are given three intervals: $I_1 = [0, x_1]$, $I = [0, x]$, and $I_2 = [0, x_2]$, where $x_2 \in [x_1 - \varepsilon_{\iou}, x_1 + \varepsilon_{\iou}]$, and $\varepsilon_{\iou} < |x_1 - x|$.
We want to find the minimum $\iou_{(2)} = \iou(I_2, I)$ as a function of $x, x_1, \varepsilon_{\iou}$.
The intersection is $I_2\cap I = [0, \min(x_2, x)]$ and the union is $I_2\cup I = [0, \max(x_2, x)]$.
Thus, the $\iou$ is:

$$\iou_{(2)} = \frac{I_2\cap I}{I_2\cup I} = \frac{\min(x_2, x)}{\max(x_2, x)}$$

We are given that $x_2 \in [x_1 - \varepsilon_{\iou}, x_1 + \varepsilon_{\iou}]$ and $\varepsilon_{\iou} < |x_1 - x|$.
We consider two cases based on the relationship between $x_1$ and $x$.

\textit{Case 1: $x_1 < x$}:
In this case, $|x_1 - x| = x - x_1$, so the condition becomes $\varepsilon_{\iou} < x - x_1$, which means $x_1 + \varepsilon_{\iou} < x$.
We have then $x_2 \le x_1 + \varepsilon_{\iou} < x$.
Therefore, $\min(x_2, x) = x_2$ and $\max(x_2, x) = x$ and
 $\iou_{(2)} = \frac{x_2}{x}$.
Assuming $x_1 - \varepsilon_{\iou} > 0$, $\min \iou_{(2)} = \frac{x_1 - \varepsilon_{\iou}}{x}$.

\textit{Case 2: $x_1 > x$}:
In this case, $|x_1 - x| = x_1 - x$, so the condition becomes $\varepsilon_{\iou} < x_1 - x$, which means $x < x_1 - \varepsilon_{\iou}$.
We have $x < x_1 - \varepsilon_{\iou} \le x_2$.
Therefore, $\min(x_2, x) = x$ and $\max(x_2, x) = x_2$ and  $\iou_{(2)} = \frac{x}{x_2}$.
Assuming $x_1 + \varepsilon_{\iou} > 0$, $\min \iou_{(2)} = \frac{x}{x_1 + \varepsilon_{\iou}}$.

Combining both cases, the minimum $\iou_{(2)}$ as a function of $x, x_1, \varepsilon_{\iou}$ is:
$$ \min \iou_{(2)} =
\begin{cases}
  \frac{x_1 - \varepsilon_{\iou}}{x} & \text{if } x_1 < x \\
  \frac{x}{x_1 + \varepsilon_{\iou}} & \text{if } x_1 > x
\end{cases}
$$

\label{eq:miniou}

\noindent\textbf{2-dim IoU}
We can extend this principle to 2-dim rectangular bounding boxes. 
In the figure \ref{fig:boundingbox}, we have chosen origin inside the intersection and we consider the $\iou$ for $x>0, y>0$,
to simplify the symbolic expressions.

Let's consider two bounding boxes, $B_1$ and $B$.
Let $B_1$ be defined by the intervals $I_{x_1} = [0, x_1]$ and $I_{y_1} = [0, y_1]$ along the x and y dimensions, respectively.
Let $B$ be defined by the intervals $I_x = [0, x]$ and $I_y = [0, y]$.
Let $B_2$ be defined by the intervals $I_{x_2} = [0, x_2]$ and $I_{y_2} = [0, y_2]$, where $x_2 \in [x_1 - \varepsilon_{\iou}, x_1 + \varepsilon_{\iou}]$ and $y_2 \in [y_1 - \varepsilon_{\iou}, y_1 + \varepsilon_{\iou}]$.

We also set the conditions $\varepsilon_{\iou} < |x_1 - x|$ and $\varepsilon_{\iou} < |y_1 - y|$.
The IoU of two bounding boxes in 2D is the ratio of the area of their intersection to the area of their union:
\begin{align*}
\iou(B_2, B) &= \iou_{(2)} = \frac{\text{Area}(B_2 \cap B)}{\text{Area}(B_2 \cup B)} \\
\text{Area}(B_2 \cup B) &= \text{Area}(B2) + \text{Area}(B) - \text{Area}(B_2 \cap B) \\
\text{Area}(B_2 \cap B) &=\min(x_2, x) \min(y_2, y) \\
 \iou_{(2)} &= \frac{\min(x_2, x) \min(y_2, y)}{x_2 y_2+x y - \min(x_2, x) \min(y_2, y)}
\end{align*}

$$ \min \iou_{(2)} =
\begin{cases}
  \frac{(x_1 - \varepsilon_{\iou})(y_1 - \varepsilon_{\iou})}{xy} & \text{if } x_1 < x, y_1 < y \\
  \frac{xy}{(x_1 + \varepsilon_{\iou})(y_1 + \varepsilon_{\iou})} & \text{if } x_1 > x, y_1 > y \\
  \frac{1}{\frac{x_1 + \varepsilon_{\iou}}{x}+\frac{y}{y_1 - \varepsilon_{\iou}}-1} & \text{if } x_1 > x, y_1 < y \\
  \frac{1}{\frac{y_1 + \varepsilon_{\iou}}{y}+\frac{x}{x_1 - \varepsilon_{\iou}}-1} & \text{if } x_1 < x, y_1 > y
\end{cases}
$$


The demonstration was performed for $x>0, y>0$. It is also applied to four quadrant by symmetry if we consider the origin in the center of $B_2\cap B$.

\end{document}